# Seeing the Big Picture: Evaluating Multimodal LLMs' Ability to Interpret and Grade Handwritten Student Work

Owen Henkel
University of Oxford
owen.henkel@gmail.com

Bill Roberts
Legible Labs

Doug Jaffe
Coherence Fund

Laurence Holt
XQ Institute

**Abstract**

Recent advances in multimodal large language models (MLLMs) raise the question of their potential for grading, analyzing, and offering feedback on handwritten student classwork. This capability would be particularly beneficial in elementary and middle-school mathematics education, where most work remains handwritten, because seeing students' full working of a problem provides valuable insights into their learning processes, but is extremely time-consuming to grade.

We present two experiments investigating MLLM performance on handwritten student mathematics classwork. Experiment A examines 288 handwritten responses from Ghanaian middle school students solving arithmetic problems with objective answers. In this context, models achieved near-human accuracy (95%, $\kappa = 0.90$) but exhibited occasional errors that human educators would be unlikely to make. Experiment B evaluates 150 mathematical illustrations from American elementary students, where the drawings are the answer to the question. These tasks lack single objective answers and require sophisticated visual interpretation as well as pedagogical judgment in order to analyze and evaluate them. We attempted to separate MLLMs' visual capabilities from their pedagogical abilities by first asking them to grade the student illustrations directly, and then by augmenting the image with a detailed human description of the illustration. We found that when the models had to analyze the student illustrations directly, they struggled, achieving only $\kappa = 0.20$ with ground truth scores, but when given human descriptions, their agreement levels improved dramatically to $\kappa = 0.47$, which was in line with human-to-human agreement levels. This gap suggests MLLMs can "see" and interpret arithmetic work relatively well, but still struggle to "see" student mathematical illustrations.

Taken together, these findings suggest that MLLMs may be capable of grading routine handwritten arithmetic work, they struggle with real-world mathematical illustrations and appear to lack both sufficient visual capabilities and the contextual, tacit knowledge necessary for the pedagogical interpretation of student mathematical thinking.

## 1 Introduction

Multimodal large language models (MLLMs) have demonstrated notable capabilities in processing visual content alongside text. Models such as GPT-4, Gemini Pro, and Claude 3.5 can analyze images, read text within them, and perform complex reasoning tasks previously challenging for automated systems. These developments raise an important question for learning analytics: can these models effectively interpret and assess handwritten student mathematical work?

The opportunity is considerable. The majority of student mathematical work, particularly in elementary and middle school, continues to be handwritten despite proliferation of digital tools [17, 21]. Teachers dedicate significant time to grading, limiting their capacity for detailed feedback [9]. More importantly, handwritten multi-step solutions reveal far more about student thinking than final answers alone—they capture problem-solving strategies, common errors, conceptual misunderstandings, and reasoning evolution [29, 30]. This rich pedagogical information has been difficult to analyze at scale due to labor-intensive manual assessment [31].

Despite growing interest in applying MLLMs to education, research on their ability to interpret authentic handwritten student mathematics remains limited. No prior work has comprehensively decomposed MLLM performance into constituent components: visual interpretation (accurately perceiving what is written), mathematical assessment (judging correctness), and pedagogical evaluation. The complexity of interpreting student work increases notably when moving from arithmetic to mathematical illustrations, such as number lines, geometric constructions, and coordinate graphs. Students may use non-standard notation, developmentally appropriate but informal representations, or creative approaches deviating from textbook methods. Interpreting such work requires not merely visual processing but extensive contextual knowledge about student development, classroom conventions, and conceptual trajectories.

Experienced educators bring tacit expertise to assessment: recognizing developmental patterns in mathematical representations; understanding recent instruction and classroom conventions; distinguishing computational errors from conceptual misunderstandings; interpreting ambiguous representations in context [15]. This knowledge, accumulated through years of experience, is rarely explicit in rubrics or guidelines [11, 33].

Our research addresses three interrelated questions:

(1) How accurately can MLLMs assess handwritten arithmetic work with objective answers?
(2) Does performance change when evaluating mathematical illustrations?
(3) Can we distinguish between MLLMs' visual and pedagogical capabilities?

To address these questions, we present two complementary experiments evaluating MLLM capabilities in interpreting handwritten student mathematical work.

Experiment A examines 288 handwritten responses to open-ended arithmetic problems from Ghanaian middle school students.

These problems involve fractions, percentages, and basic algebra where students show work and provide final answers with objective solutions. This dataset evaluates the simpler case: can models interpret relatively easy-to-interpret handwritten student work and grade it accurately?

Experiment B investigates 150 student-drawn mathematical illustrations from DrawEDU, where visual representations simultaneously constitute answers and reveal reasoning. We first evaluate MLLMs' performance grading student work providing only the image, then we provide the models the same problems with human descriptions of visual content, attempting to isolate whether limitations stem from inability to "see" versus evaluate pedagogically.

Findings reveal both promising capabilities and notable limitations. While the best model achieves near-human accuracy (95%, $\kappa = 0.90$) on arithmetic, all models show considerable degradation on illustrations. Even with visual challenges removed through human descriptions, models achieve only modest agreement with educators ($\kappa \approx 0.47$), far below practical deployment thresholds. These results suggest current MLLMs lack not just visual capabilities but situated, contextual knowledge required for meaningful interpretation of student mathematical thinking.

## 2 Related Work

### 2.1 Automated Assessment in Mathematics Education

The automation of assessment tasks in mathematics education has evolved from simple string matching to sophisticated analysis of student thinking processes. Early systems focused on multiple-choice questions or structured text inputs [18], primarily serving summative assessment needs. However, learning analytics research has increasingly recognized that assessment data can serve much richer purposes: identifying misconception patterns, tracking conceptual development, and providing actionable insights for instructional adaptation [2, 32]. The AMMORE dataset introduced by Henkel et al. [19] exemplifies this shift, providing 53,000 math open-response question-answer pairs specifically designed to capture student reasoning rather than just final answers. Another example is the work by Baral et al. [5], who developed methods for improving automated scoring of student open responses in mathematics.

Recent work has demonstrated that LLM-based approaches can effectively grade challenging student answers that rule-based classifiers fail to assess accurately [10, 28]. These systems offer the potential to analyze student work at scale, identifying not just correctness but patterns in problem-solving approaches, common error types, and conceptual gaps [10, 25]. However, most research has focused on typed responses, which represent only a fraction of authentic student mathematical production [14]. Handwritten work—with its scratch calculations, crossed-out attempts, and informal diagrams—provides a window into student thinking that typed responses cannot capture. This rich process data could enable learning analytics systems to understand not just what students know but how they think mathematically [23].

Baral et al. [4] began addressing this gap by exploring autoscoring of student responses with images in mathematics. Their work highlights an important point: handwritten mathematical work is not simply text in a different medium but a fundamentally different mode of expression that reveals cognitive processes. Students' informal sketches, their spatial arrangement of calculations, and their visual representations all provide diagnostic information that skilled educators intuitively interpret but that remains largely inaccessible to current computational methods [6]. The challenge for learning analytics is not merely technical but epistemological: how do we computationally capture the tacit knowledge that educators use when interpreting student work?

### 2.2 Multimodal Large Language Models in Education

The recognition of handwritten mathematical notation has long been a challenge for the feildof Optical Character Recognition (OCR). While recent advances in handwritten mathematical expression recognition (HMER) using transformer architectures show promise [12, 34], these approaches typically assume standard mathematical notation. Student work, however, often contains idiosyncratic representations that make perfect sense within a classroom context but confound automated systems [13]. Furthermore, building these type of custom models is time-consuming and requires significant technical expertise and resources.

Recent advancements in multimodal large language models have expanded possibilities for the interpretation of visual learning artifacts. Models such as GPT-4V [26], Claude [1], and Gemini [16] can process images alongside text, potentially enabling analysis of the full range of student mathematical productions [6]. However, their application to educational contexts reveals important gaps between generic visual understanding and the specialized interpretation required for student work.

Recent work by Hu et al. [22] on Visual Sketchpad and Zhang et al. [35] on diagram understanding suggests that current MLLMs struggle with spatial reasoning tasks that humans find intuitive. MathVista [24] and DrawEduMath [7] provide benchmarks for evaluating mathematical visual reasoning, revealing that while MLLMs can process mathematical notation and diagrams, they often lack the contextual understanding that educators bring. This limitation is particularly pronounced when interpreting children's mathematical representations, which may use non-standard notation, developmentally appropriate but mathematically informal representations, or creative problem-solving approaches that deviate from textbook methods.

This tacit knowledge is composed of extensive experience with how students at different developmental levels represent mathematical ideas, knowledge of what has been taught in specific classroom contexts, and the ability to recognize partial understanding in unconventional representations [3, 20].

### 2.3 The Challenge of Tacit Knowledge in Educational Assessment

Skilled mathematics educators bring several types of tacit knowledge to interpretation: (1) developmental understanding of how mathematical concepts typically evolve in children's thinking, (2) contextual knowledge about what specific students have previously demonstrated and what instructional approaches have been used, and (3) pattern recognition developed through years of seeing how different students express similar ideas [8, 27]. This expertise allows educators to recognize partial understanding in a hastily drawn number line or see conceptual confusion in how a student arranges their calculations.

Current approaches to automated assessment typically rely on explicit rubrics and standardized criteria. However, as Baral et al. [6] note, the gap between AI research and classroom realities partly stems from the difficulty of externalizing the tacit knowledge that expert educators apply intuitively. A student's unconventional diagram might be mathematically incorrect by strict standards yet reveal sophisticated reasoning that an experienced teacher would recognize and value. This presents a fundamental challenge for learning analytics: how can computational methods capture interpretive practices that educators themselves may struggle to articulate?

## 3 Experimental Design

This study employs a two-experiment design that investigates different aspects of MLLMs' capabilities in interpreting and grading handwritten mathematical work.

Our experimental design is grounded in the hypothesis that automated assessment of handwritten mathematics involves two distinct components: visual interpretation (accurately perceiving what is written) and grading (correctly judging the mathematical validity of what is perceived). By designing experiments that allow us to measure these components separately and in combination, we can provide a more nuanced understanding of where current MLLMs excel and where they face limitations. This decomposition is essential for understanding how to design educational technology systems that appropriately leverage MLLM capabilities while acknowledging their constraints.

We refer to these components as "vision" and "grading" to maintain clarity about what we are measuring (we avoid the term "reasoning" to prevent conflation with broader claims about model cognitive capabilities). It is important to note that in practice, the distinction between seeing correctly and grading correctly may sometimes be blurred, as interpretation of ambiguous handwriting or mathematical notation can involve elements of both visual processing and pedagogical judgment.

Experiment A focuses on the assessment of clear, handwritten numerical calculations using a novel dataset of handwritten responses from middle school students completing numerical and arithmetic operations. The high image quality and objective nature of numerical calculations in this dataset provide an ideal testing ground for understanding fundamental MLLM capabilities, establishing a baseline and allowing us to decompose performance into vision and grading components under favorable conditions.

Experiment B examines performance on mathematical illustrations and diagrams using a curated subset of the DrawEDU dataset [6]. This represents a considerably more challenging visual interpretation task that better reflects the complexity of real-world student mathematical work. We test models under two conditions: direct interpretation of images and interpretation enhanced with human-provided descriptions of the visual content. This design allows us to quantify the performance ceiling achievable when visual interpretation challenges are addressed.

Together, these experiments provide a comprehensive view of MLLM capabilities across the spectrum of handwritten mathematical work, with findings that can inform the design of educational technology systems that appropriately balance automation with human oversight.

## 4 Experiment A: Assessment of Numerical Calculations

### 4.1 Dataset and Task Design

Experiment A utilizes a novel dataset, consisting of 288 scanned handwritten responses from middle school students at Rising Academies completing numerical and arithmetic operations. These responses were collected as part of an efficacy evaluation conducted by Rising Academies, a school network based in Ghana. The problems include fractions, percentages, and basic algebra, with students required to show their work and provide final answers (see figure 1 below for an example).

We structured the evaluation by feeding models one complete worksheet at a time, with each sheet containing approximately five questions in a tabular format. Each row contained: (1) a question number, (2) the printed problem text, (3) a space for student rough work showing calculations, and (4) a designated area for the final answer.

The questions were challenging, with students achieving approximately 50% accuracy, creating a naturally balanced dataset. All responses underwent careful double-checking of the original human grading to ensure accuracy of our ground truth labels. Importantly, this dataset represents completely unseen material not present in AI training pipelines, making these first-shot evaluations particularly meaningful for assessing genuine model capabilities.

We decomposed the assessment task into two distinct components:

- **Vision Task**: Correctly identifying what numerical answer the student provided in their handwritten work;
- **Grading Task**: Correctly determining whether the student's identified answer is mathematically correct for the given problem

This separation allows us to address a fundamental question: when models fail at grading handwritten student math answers, is it because they cannot "see" the handwritten content accurately, or because they cannot correctly judge the mathematical validity of what they perceive?

### 4.2 Models and Methodology

We evaluated four state-of-the-art multimodal large language models: Claude 3.5 Sonnet, Claude 3.7, Gemini 2.5 Pro, and GPT-4.1. At the time of testing, these models represented the current frontier of multimodal AI capabilities.

For each model, we employed consistent prompting designed to elicit both visual interpretation and mathematical judgment. The prompt instructed models to: (1) identify the question number, (2) describe the student's rough work, (3) report the exact final answer written by the student, and (4) judge whether the answer is correct. This structured approach ensures consistent evaluation

Owen Henkel, Bill Roberts, Doug Jaffe, and Laurence Holt

**Figure 1: Example Student Worksheet**

```
You are being provided with a scanned copy of student math worksheet. Your task is to extract and report specific
information from each row of the worksheet

Worksheet Structure – The worksheet is organized in a table with four columns:
1 – Question Number (1-25, computer-printed); 2 – Problem (computer-printed math problem with no answer); 3 – Rough
Work (handwritten student calculations or intermediate work); 4 – Final Answer (section labeled "Final Answer:"
with the student's handwritten answer)

Your Specific Tasks – For each problem (row) on the worksheet, please:
1 – Identify the question number; 2 – Describe the student's rough work/calculations; 3 – Report the exact final
answer the student wrote; 4 – Judge whether the student's answer is correct or incorrect.

Output Format – Please structure your response as follows for each problem:
Question Number: [number]; Rough Work: [Description of the student's work/calculations]; Final Answer: [Exact
transcription of the student's final answer]; Answer Correct: A binary judgement of correctness
```

**Figure 2: Prompt for Experiment A**

| Model Description of Student Work | Model Description of Answer | GPT 4.1 Grade |
|---|---|---|
| Student wrote: 1/5 x 3/1, then 1 x 3 / 5 x 1 = 3/5, then wrote 15. | 15 | FALSE |

**Figure 3: Example of MLLM Grading Hallucination**

across models while maintaining ecological validity for educational deployment.

The prompt deliberately used minimal rubric complexity, asking only for binary correctness judgments without multi-class detailed scoring rubrics. We believe this scenario better represents how these tools would likely be deployed in authentic educational settings and provides a more realistic baseline for capabilities research in this area.

### 4.3 Results and Performance Analysis

Human-to-human agreement approached 100% for these tasks. The few initial disagreements related to edge cases—whether simplified versions of answers should be accepted, or whether to prioritize scratched-out work versus final answers—but we achieved consensus in all cases through discussion.

We evaluated performance using standard classification metrics: accuracy, precision, recall, F1 score, and Linear Weighted Kappa (LWK). The dataset was nearly balanced, with 51% of student answers being incorrect, allowing confusion matrix metrics to be interpreted directly.

Gemini 2.5 Pro notably outperforms other models, achieving 89% vision accuracy and 95% grading accuracy with $\kappa = 0.90$, indicating very high agreement with human experts. This performance advantage likely reflects its status as the largest, most recent, and most comprehensively multimodal model in our evaluation.

Table 1: Vision vs. Grading Performance Comparison

| Model | Vision | Grading | Difference |
|---|---|---|---|
| Claude 3.5 | 79% | 77% | -2% |
| Claude 3.7 | 76% | 84% | +8% |
| Gemini 2.5 Pro | 89% | 95% | +6% |
| GPT-4.1 | 86% | 79% | -7% |

Table 2: Grading Task Performance

| Model | Accuracy | LWK | F1 | Precision | Recall |
|---|---|---|---|---|---|
| Claude 3.5 | 77% | 0.54 | 0.76 | 0.82 | 0.77 |
| Claude 3.7 | 84% | 0.68 | 0.84 | 0.84 | 0.86 |
| Gemini 2.5 Pro | 95% | 0.90 | 0.95 | 0.95 | 0.95 |
| GPT-4.1 | 79% | 0.58 | 0.79 | 0.79 | 0.79 |

Table 3: Confusion Matrix Analysis

| Model | True + | False - | True - | False + |
|---|---|---|---|---|
| Claude 3.5 | 49.3% | 20.8% | 28.1% | 1.7% |
| Claude 3.7 | 49.3% | 14.2% | 34.7% | 1.7% |
| Gemini 2.5 Pro | 49.7% | 3.8% | 45.1% | 1.4% |
| GPT-4.1 | 42.0% | 12.2% | 36.8% | 9.0% |

While both Claude versions and Gemini 2.5 Pro show similar false negative rates (1.4-1.7%), the key difference lies in false positive rates. Gemini 2.5 Pro achieves only 3.8% false positives compared to 14.2-20.8% for Claude versions. This indicates that Gemini 2.5 Pro's performance advantage stems primarily from better identification of incorrect student answers—it is less likely to mistakenly accept wrong responses as correct.

### 4.4 Vision-Grading Decomposition Analysis

The relationship between vision and grading performance reveals several interesting patterns across models: These vision accuracy scores should, in principle, establish a ceiling for grading performance—if a model cannot accurately see what a student wrote, it presumably cannot judge correctness and would resort to guessing. Given our balanced dataset (approximately 50% correct/incorrect), random guessing would degrade performance toward 50% accuracy.

Claude 3.7 and Gemini 2.5 Pro demonstrate grading performance that exceeds their vision accuracy by 8 and 6 percentage points respectively, suggesting these models may use mathematical context to compensate for imperfect visual interpretation. Conversely, GPT-4.1 shows a 7-percentage-point drop from vision to grading, with notably higher false negative rates, indicating a tendency to incorrectly classify correct student responses as wrong even when visual content is accurately perceived.

Despite these nuanced interactions, vision capability remains a strong predictor of overall performance, with the 13-percentage-point spread between weakest and strongest vision performers largely predicting their relative ranking on grading tasks.

Perhaps most puzzling are failure modes that seem fundamentally different from human reasoning patterns (see figure 3 for an example). The student's intermediate calculation contains errors but arrives at the correct answer, yet the model penalizes the faulty mathematical process over the correct outcome. This type of error—rejecting correct final answers due to flawed intermediate steps—highlights the unpredictable nature of MLLM reasoning, even in seemingly objective tasks.

## 5 Experiment B: Assessment of Mathematical Illustrations

### 5.1 Dataset and Task Design

Experiment B utilizes a curated subset of 150 examples from the DrawEDU dataset [6], featuring student-drawn responses to mathematical problems. This dataset presents mathematical illustrations, diagrams, and visual representations including number lines, geometric shapes, coordinate planes, and other spatial mathematical content requiring visual interpretation beyond simple character recognition. Unlike the objective numerical calculations in Experiment A, interpreting these mathematical illustrations involves pedagogical judgment. Students may use non-standard notation, partially correct approaches, or ambiguous visual representations that require interpretation. Common challenges include cases where students provide correct formula work but incorrect accompanying illustrations, or vice versa, creating genuine ambiguity about appropriate grading under a binary rubric.

We designed two experimental conditions to isolate the impact of visual interpretation challenges:

A) Round to the nearest ten.

26 ≈ \_\_\_\_\_\_\_\_

B) Use the number line to model your thinking.

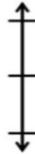

Example A : Correct     Example B : Incorrect

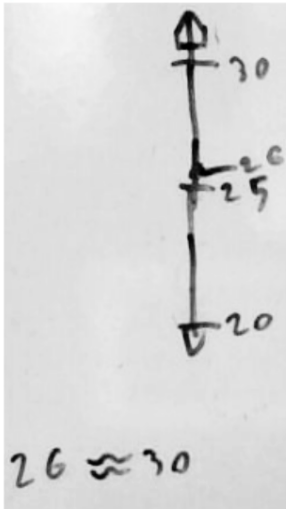 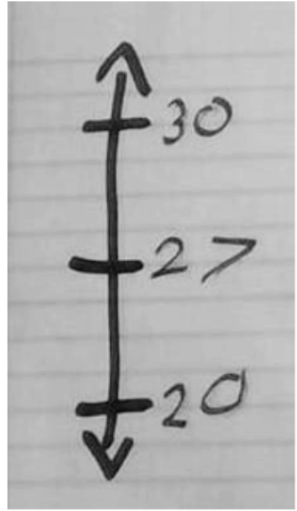

Figure 4: Example Student Worksheet

- **Task 1 (Model-Only)**: Models view only the image and determine whether the student response is correct or incorrect
- **Task 2 (Human-Enhanced)**: Models receive high-quality, objective human descriptions of the visual content, alongside the image, and determine whether the student response is correct or incorrect

This two-condition design allows us to quantify the performance ceiling achievable when visual interpretation challenges are minimized, providing guidance on when human-in-the-loop approaches are necessary.

### 5.2 Human Inter-Rater Agreement

Establishing ground truth for mathematical illustrations required careful procedures. Every item was independently rated by two expert mathematics teachers, achieving initial agreement of approximately 76% with a LWK of 0.45. A third rater adjudicated disagreements through careful review, achieving consensus in nearly all cases.

Approximately 10-20 items out of 150 presented reasonable ambiguity where expert educators could legitimately disagree about correctness. This inherent ambiguity creates a performance ceiling below perfect agreement, meaning that kappa scores around 0.45-0.50 may represent near-optimal performance for this task rather than indicating poor model capabilities.

For Task 2, human descriptions were created by the same annotators who objectively described the visual content without making judgments about correctness. These descriptions included details about shapes drawn, numbers written, positions of elements, and spatial relationships, providing models with accurate visual information while still requiring them to make the mathematical assessment.

### 5.3 Models and Methodology

We evaluated the same four models using consistent prompting strategies. The rubric instructed models to treat student responses as correct if they showed substantial correctness or clear evidence of the right answer, but incorrect if ambiguous, unclear, or demonstrably wrong. We deliberately maintained minimal rubric complexity to preserve generalizability across different educational contexts.

### 5.4 Results

Table 4: Task 1 Performance (Image Only)

| Model | Accuracy | Kappa Score | Precision | Recall | F1 |
| --- | --- | --- | --- | --- | --- |
| Claude 3.5 | 64% | 0.23 | 0.65 | 0.64 | 0.64 |
| Claude 3.7 | 63% | 0.15 | 0.62 | 0.63 | 0.62 |
| GPT-4.1 | 61% | 0.23 | 0.66 | 0.61 | 0.61 |
| Gemini 2.5 | 64% | 0.38 | 0.72 | 0.71 | 0.71 |

Table 5: Task 2 Performance (With Human Descriptions)

| Model | Accuracy | Kappa Score | Precision | Recall | F1 |
| --- | --- | --- | --- | --- | --- |
| Claude 3.5 | 74% | 0.46 | 0.76 | 0.74 | 0.74 |
| Claude 3.7 | 75% | 0.47 | 0.76 | 0.75 | 0.75 |
| GPT-4.1 | 73% | 0.47 | 0.80 | 0.72 | 0.73 |
| Gemini 2.5 | 72% | 0.43 | 0.75 | 0.72 | 0.72 |

Table 6: Performance Improvement Summary

| Model | Task 1 (Direct) | Task 2 (Human+) | Improvement |
| --- | --- | --- | --- |
| Claude 3.5 | $\kappa$=0.23, 64% | $\kappa$=0.46, 74% | +0.23 $\kappa$, +10% |
| Claude 3.7 | $\kappa$=0.15, 63% | $\kappa$=0.47, 75% | +0.32 $\kappa$, +12% |
| GPT-4.1 | $\kappa$=0.23, 61% | $\kappa$=0.47, 73% | +0.24 $\kappa$, +12% |
| Gemini 2.5 | $\kappa$=0.38, 64% | $\kappa$=0.43, 72% | +0.05 $\kappa$, +8% |

## 5.5 Analysis

All models show substantial performance gains when provided with human descriptions, with Claude 3.7 showing the largest improvement ($\kappa$ gain of +0.32). This substantial improvement confirms that visual interpretation remains a major bottleneck for current MLLMs when processing complex mathematical illustrations. The performance gains from human descriptions demonstrate that models still face significant limitations in interpreting student-drawn mathematical illustrations.

Interestingly, all models converge to nearly identical performance levels in Task 2 ($\kappa \approx$ 0.43-0.47, accuracy$\approx$72-75%), approaching levels similar to human inter-rater agreement before calibration. This convergence raises an important interpretive challenge: when model performance reaches $\kappa \approx$0.47—comparable to the initial human inter-rater agreement of $\kappa$=0.45—it becomes difficult to determine how much of the remaining performance gap reflects inherent subjectivity in the assessment task versus the models' inability to interpret student answers as an educator would.

However, the $\kappa \approx$0.47 ceiling is clearly insufficient for deployment in real-world educational settings.

## 6 Discussion

### 6.1 Performance Patterns and the Tacit Knowledge Gap

Our experiments reveal a notable divide in MLLM capabilities between routine arithmetic and mathematical illustrations. While models achieve near-human performance (95% accuracy, $\kappa$=0.90) on numerical calculations with objective answers, performance drops to $\kappa \approx$0.47 on illustrations even with human visual descriptions providing optimal visual information.

The vision-grading decomposition reveals unexpected patterns. Some models achieve higher grading accuracy than vision accuracy, suggesting they compensate for imperfect visual interpretation through mathematical context. While potentially useful for identifying student understanding despite messy handwriting, this raises concerns about false positives. More troubling are non-human error patterns—such as rejecting correct answers due to untidy intermediate work—highlighting fundamental differences in how models versus educators interpret student thinking.

The substantial performance improvements with human descriptions ($\kappa$ improvements of 0.23-0.32) confirm visual interpretation as a major bottleneck. However, the plateau at $\kappa \approx$0.47 reveals a deeper challenge: current MLLMs with minimal prompting cannot replicate educators' tacit knowledge. This includes developmental understanding of how mathematical representations evolve across grade levels, awareness of classroom-specific methods and conventions, and the ability to recognize sophisticated reasoning in imprecise drawings. A hastily drawn number line or unconventional diagram might reveal deep understanding to an experienced teacher while confounding an MLLM. This gap reflects not just technical limitations but the fundamental challenge of computationally capturing interpretive practices that educators themselves struggle to articulate.

### 6.2 Implications for System Designers and Practitioners

These findings delineate clear boundaries for learning analytics applications. For routine arithmetic with objective answers, MLLMs could support automated data collection—identifying struggling students, tracking accuracy trends, and flagging unusual patterns for review. This enables more frequent formative assessment without increasing teacher workload. However, mathematical illustrations require human-in-the-loop approaches where MLLMs serve as intelligent filters, processing large volumes to identify cases needing expert interpretation.

Implementation strategies vary by use case. Near-term deployments should focus on hybrid systems: MLLMs handle initial screening of arithmetic work while flagging edge cases for educator review; for illustrations, models could pre-sort student work by confidence levels, allowing teachers to focus attention where human judgment is most critical. System designers should prioritize transparency, showing educators not just grades but also what features the model attended to, enabling teachers to quickly identify and correct systematic errors. Professional development will be essential to help educators understand both the capabilities and limitations of these tools, ensuring they augment rather than replace pedagogical expertise.

### 6.3 Limitations

Our binary rubric structure and minimal prompting approach, while establishing baseline capabilities, likely underestimates what might be achieved with careful prompt engineering. Future work should explore whether sophisticated prompting strategies could elicit more nuanced evaluation. However, we believe our conservative approach better reflects the practical reality of deployment in educational settings, where extensive prompt engineering for every assessment context may not be feasible. The focus on specific grade levels and mathematical domains limits generalizability, though consistent performance patterns across different models suggest these limitations reflect fundamental challenges in computational interpretation of student thinking rather than model-specific weaknesses.

## 7 Conclusion

This study evaluates MLLMs' capabilities in interpreting handwritten student mathematical work—a critical challenge for learning analytics seeking to understand student thinking through authentic classroom artifacts. Our decomposition of performance into visual interpretation and assessment components reveals both promising capabilities and important limitations.

Current MLLMs achieve near-human performance on routine arithmetic but struggle significantly with mathematical illustrations. Even when human descriptions eliminate visual processing challenges, models plateau at moderate agreement levels ($\kappa \approx$0.47), comparable to initial human inter-rater agreement but insufficient for autonomous deployment. This ceiling appears to reflect both inherent task ambiguity and models' lack of the situated, contextual knowledge that makes expert interpretation possible.

Effective deployment requires careful consideration of automation boundaries. The goal should be amplifying educator expertise through computational methods for routine analysis while preserving human insight for complex interpretive work essential to understanding mathematical learning. As these technologies develop, maintaining focus on authentic student understanding rather than assessment accuracy alone will be critical for realizing their educational potential.

The path forward requires both immediate practical steps and longer-term research initiatives. In the near term, we need hybrid systems that transparently integrate MLLM capabilities for arithmetic assessment while maintaining educator oversight for ambiguous cases. Professional development programs must help teachers effectively collaborate with AI tools, understanding when to trust automated assessments and when human judgment remains essential. Looking ahead, the research community must develop architectures that better integrate visual and conceptual reasoning, potentially through specialized preprocessing pipelines or multi-stage interpretation systems. Equally important is the long-term challenge of capturing and operationalizing educators' tacit knowledge—requiring interdisciplinary collaboration to document expert interpretation processes, build annotated corpora with rich contextual metadata, and design interactive systems where educational expertise and computational power reinforce each other rather than compete.

## Digital Appendix

All analysis code, datasets, experimental materials, and supplementary documentation can be found in the study's GitHub repository: https://github.com/owenhenkel/seeing-the-big-picture